\documentclass[10pt,twocolumn,letterpaper]{article}

\usepackage{cvpr}              

\usepackage[utf8]{inputenc}
\usepackage[T1]{fontenc}
\usepackage{amsfonts}
\usepackage{amsmath}
\usepackage{bm}
\usepackage{booktabs}
\usepackage{tabularx}
\newcolumntype{C}{>{\centering\arraybackslash}X}
\usepackage{graphicx}
\usepackage{microtype}
\usepackage{nicefrac}
\usepackage{pifont}
\usepackage{url}
\usepackage{xcolor}
\usepackage{xspace}
\usepackage{multirow}

\newcommand{\cmark}{\ding{51}}%
\newcommand{\xmark}{\ding{55}}%

\newcommand{\method}{Flash3D\xspace}

\newcommand{\x}{\boldsymbol{x}}

\newcommand{\bu}{\boldsymbol{u}}
\newcommand{\bmu}{\boldsymbol{\mu}}
\newcommand{\bnu}{\boldsymbol{\nu}}

\definecolor{cvprblue}{rgb}{0.21,0.49,0.74}
\usepackage[pagebackref,breaklinks,colorlinks,citecolor=cvprblue]{hyperref}

\def\paperID{413} 
\def\confName{3DV\xspace}
\def\confYear{2025\xspace}

\title{\method:
Feed-Forward Generalisable 3D \\ Scene Reconstruction from a Single Image}

\author{ Stanislaw Szymanowicz$^1$\thanks{denotes equal contribution} \quad Eldar Insafutdinov$^{1*}$ \quad Chuanxia Zheng$^{1*}$ \\
 Dylan Campbell$^2$ \quad João F. Henriques$^1$ \quad Christian Rupprecht$^1$ \quad Andrea Vedaldi$^1$ \\ [0.3em]
$^1$VGG, University of Oxford \quad
$^2$Australian National University \\
{\tt\small \{stan,eldar,cxzheng,joao,chrisr,vedaldi\}@robots.ox.ac.uk dylan.campbell@anu.edu.au}
}

\begin{document}
\maketitle
\vspace{-0.5em}

\begin{abstract}
\vspace{-5pt}
We propose \method, a method for scene reconstruction and novel view synthesis from a single image which is both very generalisable and efficient.
For generalisability, we start from a `foundation' model for monocular depth estimation and extend it to a full 3D shape and appearance reconstructor.
For efficiency, we base this extension on feed-forward Gaussian Splatting.
Specifically, we predict a first layer of 3D Gaussians at the predicted depth, and then add additional layers of Gaussians that are offset in space, allowing the model to complete the reconstruction behind occlusions and truncations.
\method is very efficient, trainable on a single GPU in a day, and thus accessible to most researchers.
It achieves state-of-the-art results when trained and tested on RealEstate10k.
When transferred to \emph{unseen} datasets like NYU it outperforms competitors by a large margin.
More impressively, when transferred to KITTI, \method achieves better PSNR than methods trained specifically on that dataset.
In some instances, it even outperforms recent methods that use multiple views as input.
Code, models, demo, and more results are available at \href{https://www.robots.ox.ac.uk/~vgg/research/flash3d/}{https://www.robots.ox.ac.uk/\textasciitilde vgg/research/flash3d/}.
\end{abstract}
\section{Introduction}%
\label{sec:intro}

We consider the problem of reconstructing photorealistic 3D scenes from a single image in just one forward pass of a network.
This is a challenging task because scenes are complex and monocular reconstruction is ill-posed.
Unambiguous geometric cues, such as triangulation, are unavailable in the monocular setting, and there is no direct evidence of the occluded parts of the scene.

This problem is closely related to monocular depth estimation~\cite{shao23nddepth:,saxena23monocular,lasinger19towards,ocal20realmonodepth:,freeman19learning,cao18estimating,godard16unsupervised,laina16deeper,birchfield98depth,buxton83monocular,yang24depth,piccinelli24unidepth:}, which is a mature area.
It is now possible to accurately estimate metric depth with excellent cross-domain generalisation~\cite{yang24depth,piccinelli24unidepth:,yin23metric3d:}.
However, while depth estimators predict the 3D shape of the nearest visible surfaces, they do not provide any \emph{appearance} information, nor an estimate of the occluded or out-of-frame parts of the scene.
Depth alone is insufficient to accurately solve tasks such as \emph{novel view synthesis} (NVS), which additionally require modelling unseen regions and view-dependent appearance.

While methods for monocular scene reconstruction exist~\cite{tucker20single-view,li21mine:,wimbauer23behind}, they mostly operate in a `closed-world' setting where they are trained anew for each considered dataset.
In contrast, modern depth predictors generalise well to new datasets at inference time.
Furthermore, current monocular scene reconstructors are often slow or incur a high computational memory cost due to volumetric rendering~\cite{li21mine:} and implicit representations~\cite{yu21pixelnerf:}.

Very recently, \citet{szymanowicz24splatter} introduced the Splatter Image (SI), a method for fast monocular reconstruction of individual objects that builds upon the success of Gaussian Splatting~\cite{kerbl233d-gaussian}.
The approach is simple: predict the parameters of a coloured 3D Gaussian for each input image pixel using a standard image-to-image neural network architecture.
The resulting Gaussian mixture was shown to reconstruct objects well, including unobserved surfaces.
In part, this was due to the fact that SI is able to use some of the ``background pixels'' to model the occluded parts of the object.
However, in scene reconstruction, there is not such a reservoir of background pixels, which poses a challenge for the method.
In contrast, pixelSplat~\cite{charatan23pixelsplat:}, MVSplat~\cite{chen24mvsplat:}, latentSplat~\cite{wewer24latentsplat:} and GS-LRM~\cite{zhang24gs-lrm:}, which share a similar design, were designed for scene reconstruction; however, they address the binocular reconstruction problem, requiring \emph{two} images of the scene captured from different \textit{known} viewpoints.
We instead consider the more challenging monocular setting, since it is more generally applicable and does not require camera extrinsics, which is a challenging research problem on its own~\cite{wang23dust3r:,wang23posediffusion:,zhang24cameras}.

\begin{figure*}[tb!]
    \centering
    \includegraphics[width=0.9\textwidth]{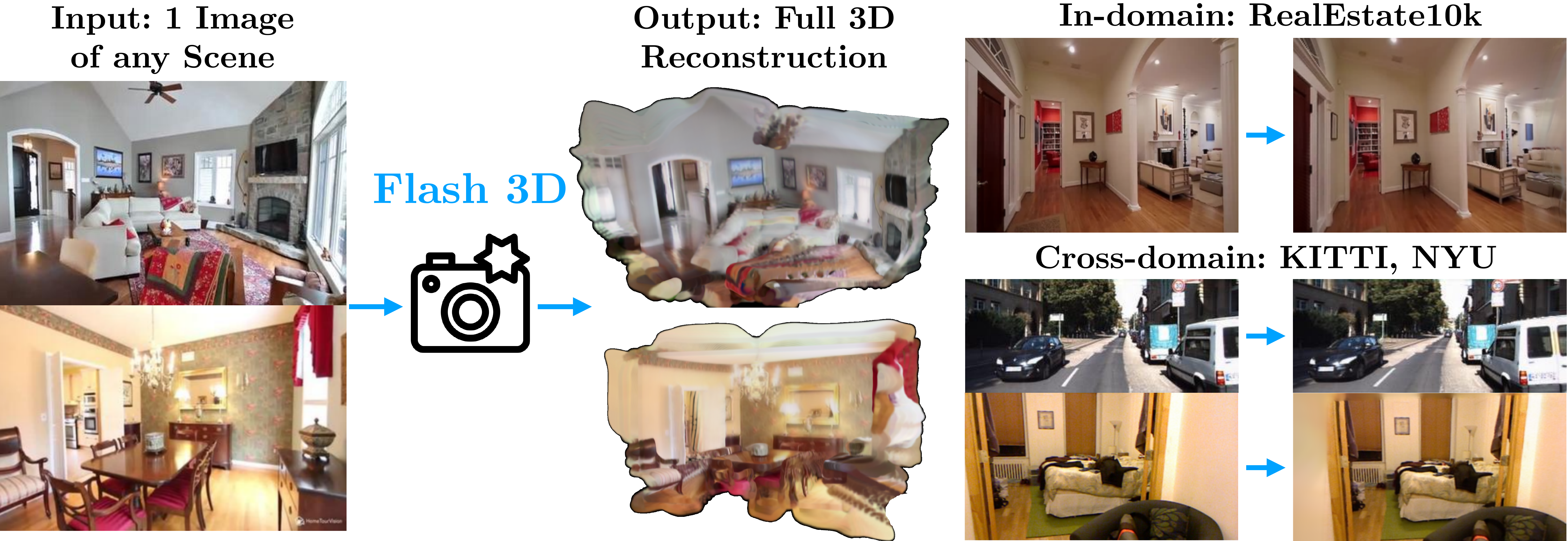}
    \caption{\textbf{\method} reconstructs the 3D (\textit{not} 2.5D) scene structure and appearance from just a single image `in a flash', enabling accurate novel view synthesis.
    Trained on just one dataset, it generalises to new, different datasets and unknown scenes.
    }
    \label{fig:teaser}
\end{figure*}

In this work, we introduce a new, simple, efficient and performant approach for monocular scene reconstruction called \emph{\method}.
This is based on two key ideas.
First, we address the issue of \emph{generalisation} which limits current feed-forward monocular scene reconstructors.
The aim is for \method to work on any scene, not just on scenes similar to the ones in the training set.
Analogous open-ended models are often called foundation models and require massive training datasets and computational resources unavailable to most research groups.
A similar problem exists in 3D object reconstruction and generation~\cite{melas-kyriazi23realfusion,shi24mvdream:,li24instant3d:,melas-kyriazi24im-3d,liu23zero-1-to-3:,zheng24free3d}, where it is addressed by extending to 3D an existing foundation 2D image or video generator~\cite{rombach21geometry-free,podell23sdxl:,blattmann23stable,dai23emu:,girdhar23emu-video:}.
Here, we posit that scene reconstruction can also benefit from building on an existing foundation model, but opt for a monocular depth predictor as a more natural choice.
We show, in particular, that by building on a high-quality depth predictor~\cite{piccinelli24unidepth:}, we can achieve excellent generalisation to new datasets, to the point that our 3D reconstructions are more accurate than those of models trained specifically on those test domains.

Second, we improve feed-forward per-pixel Gaussian splatting for monocular scene reconstruction.
As noted, applied to single objects, a per-pixel reconstructor can use the reservoir of background pixels to model the hidden parts of the object, which is not possible when reconstructing a full scene.
Our solution is to predict multiple Gaussians per pixel, where only the first Gaussian along each ray is encouraged to conform to the depth estimate, and thus model the visible part of the scene.
This is analogous to a layered representation~\cite{adelson95layered,baker98a-layered,shade98layered,tulsiani18layer-structured,tucker20single-view,li21mine:} and multi-Gaussian sampling in pixelSplat~\cite{charatan23pixelsplat:}.
However, in our case Gaussians are deterministic, not limited to specific depth ranges, and the model is free to shift Gaussians off the ray to model occluded or truncated parts of the scene.

Overall, \method is a simple and highly performant monocular scene reconstruction pipeline.
Empirically, we find that \method can
(a)~render high-quality images of the reconstructed 3D scene,
(b)~operate on a wide range of scenes, both indoor and outdoor; and
(c)~reconstruct occluded regions, which would not be possible with depth estimation alone or with na{\"\i}ve extensions of it.
\method achieves state-of-the-art novel view synthesis accuracy in all metrics on  RealEstate10K~\cite{tinghui18stereo}.
More impressively, the same frozen model also achieves state-of-the-art accuracy when transferred to NYU~\cite{silberman12indoor} and KITTI~\cite{geiger13vision} (in PSNR).
Furthermore, in an extrapolation setting, our reconstructions can even be more accurate than those of binocular methods like pixelSplat \cite{charatan23pixelsplat:} and latentSplat \cite{wewer24latentsplat:} that use two images of the scene instead of one, and are thus at a significant advantage.

In addition to the quality of the reconstructions, shown in \cref{fig:teaser}, \method is very efficient to evaluate and, most importantly, to train.
For instance, we use 1/64\textsuperscript{th} of the GPU resources of prior works such as MINE \cite{li21mine:}.
By achieving state-of-the-art results while using modest computational resources for training, this opens the research area to a wider range of researchers.

\section{Related Work}
\label{sec:related_work}
\begin{figure*}[tb!]
    \centering
    \includegraphics[width=\textwidth]{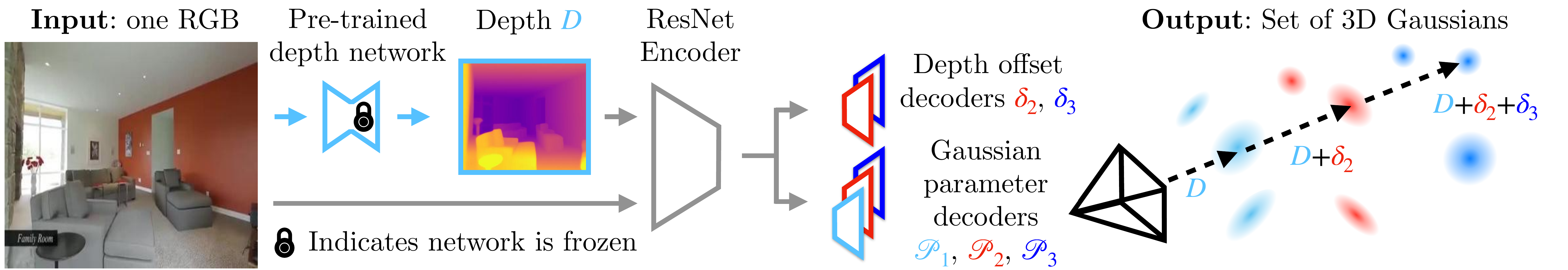}
    \caption{\textbf{Overview of \method.}
    %
    Given a single image $I$ as input, \method first estimates the metric depth $D$ using a frozen off-the-shelf network~\cite{piccinelli24unidepth:}.
    Then, a ResNet50-like encoder--decoder network predicts a set of  
    shape and appearance parameters $\mathcal{P}$ of $K$ layers of Gaussians  
    for every pixel $\bu$, allowing unobserved and occluded surfaces to be modelled.
    From these predicted components, the depth can be obtained by summing the predicted (positive) offsets $\delta_{i}$ with the predicted monocular depth $D$, allowing the mean vector for every layer of Gaussians to be computed.
    This strategy ensures that the layers are depth-ordered, encouraging the network to model occluded surfaces.
    }
    \label{fig:framework}
\end{figure*}

\paragraph{Monocular feed-forward reconstruction.}

Like our approach, monocular feed-forward reconstructors work by passing a single image of the scene through a neural network to output a 3D reconstruction directly.
For \emph{scenes}, the works of~\cite{tulsiani18layer-structured,tucker20single-view,tulsiani17multi-view} and MINE~\cite{li21mine:} do so by predicting multi-plane images~\cite{tinghui18stereo}, while BDS \cite{wimbauer23behind} uses neural radiance fields \cite{sitzmann19scene,mildenhall20nerf:}.
Our method outperforms them in terms of speed and generalisation.
Like our work, SynSin~\cite{wiles20synsin:} uses a monocular depth predictor to reconstruct a scene; however, its reconstructions are incomplete and require a rendering network to improve the final novel views.
In contrast, our approach outputs a high-quality 3D reconstruction which can be directly rendered with Gaussian Splatting~\cite{kerbl233d-gaussian}.
For \emph{objects}, a notable example is Large Reconstruction Model (LRM)~\cite{hong24lrm:}, which obtains high-quality monocular reconstruction with a very large, and costly to train, model.
The most related work is Splatter Image~\cite{szymanowicz24splatter} that uses Gaussian Splatting~\cite{kerbl233d-gaussian} for efficiency.
Our approach also uses Gaussian Splatting as a representation, but does so for scenes rather than objects, which presents different challenges.

\paragraph{Few-view feed-forward reconstruction.}

A less challenging but still important case is few-view feed-forward reconstruction, where reconstruction requires two or more images.
Early examples used NeRFs~\cite{mildenhall20nerf:} as 3D representation of \emph{objects}~\cite{chibane21stereo,henzler21unsupervised,johari22geonerf:,liu22neural,yu21pixelnerf:,reizenstein21common,wang21ibrnet:} and scenes~\cite{chibane21stereo,du23learning,xu23murf:}.
These methods implicitly learn to match points between views; the works of~\cite{chen21mvsnerf:,yuedong23explicit} make point matching more explicit.
While many few-view reconstructors estimate the 3D shape of the object as an opacity field, an alternative is to
directly predict new views~\cite{miyato23gta:,sajjadi21scene,suhail21light,suhail22generalizable} of \emph{scenes} with no explicit volumetric reconstruction, a concept pioneered by LFNs~\cite{sitzmann21light}.
Other works use instead multi-plane images from narrow baseline stereo pairs~\cite{tinghui18stereo,srinivasan19pushing} and few views~\cite{mildenhall19local,kar17learning}.
More related to our approach, pixelSplat~\cite{charatan23pixelsplat:}, latentSplat~\cite{wewer24latentsplat:} and MVSplat~\cite{chen24mvsplat:} reconstruct scenes from a pair of images.
They utilise cross-view attention to share information and predict Gaussian mixtures to represent the scene geometry.
Other very recent approaches~\cite{tang24lgm:,xu24grm:,zhang24gs-lrm:} combine LRM and Gaussian Splatting for reconstruction from a small number of images.
We address \emph{monocular} reconstruction instead, which is a much harder problem due to lack of geometric cues from triangulation.

\paragraph{Iterative reconstruction.}


Iterative or optimisation-based methods reconstruct from one or more images by iteratively fitting a 3D model to them.
Due to their iterative nature, and the need to render the 3D model to fit it to the data, they are generally much slower than feed-forward approaches.
%
DietNeRF~\cite{jain21putting} regularises reconstruction using language models, RegNeRF~\cite{niemeyer22regnerf:} and RefNeRF~\cite{verbin22ref-nerf:} use handcrafted regularisers, and SinNeRF~\cite{xu22sinnerf:} uses monocular depth.
RealFusion~\cite{melas-kyriazi23realfusion} uses an image diffusion model as a prior for monocular reconstruction based on slow score distillation sampling iterations~\cite{poole23dreamfusion:}.
Numerous follow-ups took a similar path~\cite{tang23make-it-3d:,wynn23diffusionerf:}.
Convergence speed and robustness can be improved by using multi-view aware generators~\cite{watson23novel,liu23zero-1-to-3:,melas-kyriazi24im-3d,gu23nerfdiff:,tseng23consistent,zheng24free3d}.
Approaches like Viewset Diffusion~\cite{szymanowicz23viewset} and RenderDiffusion~\cite{anciukevicius22renderdiffusion:} fuse 3D reconstruction with diffusion-based generation, which can reduce but not eliminate the cost of iterative generation. 
More recently, ZeroNVS~\cite{sargent23zeronvs:}, ReconFusion~\cite{wu2024reconfusion} and Cat3D~\cite{gao2024cat3d} utilize the large-scale diffusion model for single view or sparse view 3D reconstruction and NVS.
However, they focus on ``realistic'' view generation, instead of ``accurate'' 3D reconstruction, and they still need the per-scene optimisation to achieve consistent 3D structure, which is relative expensive and slow.
In contrast, our approach is feed-forward and therefore significantly faster, close to real-time (10fps).
%
Some approaches generate novel views in a feed-forward manner, but iteratively and autoregressively, one view at a time.
Examples include PixelSynth \cite{rockwell21pixelsynth:}, GeNVS~\cite{chan23generative}, and Text2Room~\cite{hollein23text2room:}.
In contrast, we generate the final 3D reconstruction in a single feed-forward pass.

\paragraph{Monocular depth prediction.}

Our method is based on monocular depth estimation \cite{buxton83monocular,
birchfield98depth,
eigen14depth,
laina16deeper,
godard16unsupervised,
zhou17unsupervised,
cao18estimating,
godard19digging,
freeman19learning,
lasinger19towards,
ocal20realmonodepth:,
ranftl21vision,
shao23nddepth:,
saxena23monocular,
yang24depth}, where metric or relative depth is predicted for every image pixel for a given image.
By learning visual depth cues from large datasets,
these approaches have demonstrated high accuracy and the capacity to generalise across datasets.
While our method is agnostic to the depth predictor used, we use one of the state-of-the-art metric depth estimators, UniDepth~\cite{piccinelli24unidepth:}, for our experiments.
\section{Method}%
\label{s:method}

Let $I \in \mathbb{R}^{3\times H\times W}$ be an RGB image of a scene.
Our goal is to learn a neural network $\Phi$ that takes as input $I$ and predicts a representation $\mathcal{G} = \Phi(I)$ of the 3D content of the scene, both in terms of 3D geometry and photometry.
We first discuss the background and baseline model in \cref{s:background}, introduce our layered multi-Gaussian predictor and discuss the use of monocular depth prediction as a prior in \cref{s:prior}.

\subsection{Background: Scene reconstruction from a single image}%
\label{s:background}

\paragraph{Representation: scenes as sets of 3D Gaussians.}

The scene representation
$
\mathcal{G} = \{(\sigma_i,\bmu_i,\Sigma_i,c_i)\}_{i=1}^G
$
is a set of 3D Gaussians~\cite{kerbl233d-gaussian}, where $\sigma_i \in [0, 1)$ is the opacity, $\bmu_i\in\mathbb{R}^3$ is the mean, $\Sigma_i\in\mathbb{R}^{3\times 3}$ is the covariance matrix, and 
$
c_i : \mathbb{S}^2 \rightarrow \mathbb{R}^3
$
is the radiance function (directional colour) of each component.
Let
$
g_i(\x)
=
\exp\left(
  -\frac{1}{2}(\x-\bmu_i)^\top\Sigma_i^{-1}(\x-\bmu_i)
\right)
$
be the corresponding (un-normalised) Gaussian function.
The colours of the Gaussians are generally represented using spherical harmonics, so that
$
[c_i(\bnu)]_j = \sum_{l=0}^L \sum_{m=-l}^l c_{ijlm} Y_{lm}(\bnu),
$
where $\bnu \in \mathbb{S}^2$ is a view direction and $Y_{lm}$ are the spherical harmonics of various orders $m$ and degrees $l$.
The Gaussian mixture $\mathcal{G}$ defines the opacity and colour functions of a radiance field:
$
\sigma(\x) = \sum_{i=1}^G \sigma_i g_i(\x)
$,
$
c(\x,\bnu)
= 
{
  \sum_{i=1}^G c_i(\bnu) \sigma_i g_i(\x)
} / {
  \sum_{i=1}^G \sigma_i g_i(\x)
}
$,
where $\sigma(\x)$ is the opacity at 3D location $\x\in\mathbb{R}^3$ and $c(\x,\bnu)$ is the radiance at $\x$ in direction $\bnu\in\mathbb{S}^2$ towards the camera.

The field is rendered into an image $J$ by integrating the radiance along the line of sight using the \emph{emission--absorption}~\cite{DBLP:journals/tvcg/Max95a} equation
$
J(\bu)
=
\int_{0}^\infty
c(\x_t,\bnu) \sigma(\x_t)
\exp{(-\int_0^t \sigma(\x_\tau) d\tau)}
dt
$,
where $\x_t = \x_0 - t \bnu$ is the ray originating at the camera centre $\x_0$ and propagating towards the pixel $\bu$ in the direction $-\bnu$.
The key contribution of Gaussian Splatting~\cite{kerbl233d-gaussian} is to 
approximate this integral very efficiently, implementing a differentiable rendering function $\hat J = \operatorname{Rend}(\mathcal{G}, \pi)$ which takes as input the Gaussian mixture $\mathcal{G}$ and viewpoint $\pi$ and returns an estimate $\hat J$ of the corresponding image.

\paragraph{Monocular reconstruction.}

Following~\cite{szymanowicz24splatter}, the output $\Phi(I)\in\mathbb{R}^{C \times H\times W}$
of the neural network is a tensor that specifies, for each pixel $\bu=(u_x,u_y,1)$, the parameters of a coloured Gaussian, consisting of the opacity $\sigma$, the depth $d \in \mathbb{R}_+$, the offsets $\Delta \in \mathbb{R}^3$, the covariance $\Sigma \in \mathbb{R}^{3\times 3}$ expressed as rotation and scale (seven parameters, using quaternions for rotation), and the parameters of the colour model $c \in \mathbb{R}^{3(L+1)^2}$ where $L$ is the order of the spherical harmonics.
The mean of the Gaussian is then given by
$
\bmu = K^{-1} \bu d + \Delta,
$
where $K=\operatorname{diag}(f,f,1)\in\mathbb{R}^{3\times 3}$ is the camera calibration matrix and $f$ its focal length.
Hence, there are $C = 1 + 1 + 3 + 7 + 3(L+1)^2 = 12 + 3(L+1)^2$
parameters predicted for each pixel.
The model $\Phi$ is trained using triplets $(I,J,\pi)$ where $I$ is an input image, $J$ is a target image, and $\pi$ is the relative camera pose.
To learn the network parameters, one simply minimises the rendering loss
$
\mathcal{L}(\mathcal{G},\pi,J) = \|\operatorname{Rend}(\mathcal{G}, \pi) - J\|
$.

\subsection{Monocular feed-forward multi-Gaussians}%
\label{s:prior}


For generalisation, we propose to build \method on a high-quality pre-trained model trained on a large amount of data.
Specifically, given the similarities between monocular scene reconstruction and monocular depth estimation, we use an off-the-shelf monocular depth predictor $\Psi$.
This model takes as input an image $I$ and returns a depth map $D = \Psi(I)$, where $D \in \mathbb{R}_+^{H \times W}$ is a matrix of depth values, as explained next.

\paragraph{Baseline architecture.}

Given an image $I$ and estimated depth map $D$, our baseline model consists of an additional network $\Phi(I, D)$ that takes as input the image and the depth map and returns the required per-pixel Gaussian parameters.
In more detail, for each pixel $\bu$, the entry $[\Phi(I,D)]_{\bu} = (\sigma, \Delta, s, \theta, c)$ consists of the opacity $\sigma \in \mathbb{R}_{+}$, the displacement
$\Delta \in \mathbb{R}^3$,
the scale $s \in \mathbb{R}^3$, the quaternion $\theta \in \mathbb{R}^4$ parametrising the rotation $R(\theta)$, and the colour parameters $c$.
The covariance of each Gaussian is given by
$
\Sigma =
R(\theta)^\mathsf{T} \operatorname{diag}(s) R(\theta)
$
and the mean is given \cite{szymanowicz24splatter} by
$
\bmu = (u_x \, d / f, \, u_y \, d / f, \, d) + \Delta,
$
where $f$ is the focal length of the camera (either known or also estimated by $\Psi$) and the depth $d = D(\bu)$ is from the depth map.
The network $\Phi$ is a U-Net~\cite{ronneberger15u-net:} utilising ResNet blocks~\cite{he16deep} for encoding and decoding, denoted $\Phi_{\text{enc}}$ and $\Phi_{\text{dec}}$ respectively.
The decoder network thus outputs a tensor
$
\Phi_{\text{dec}}(\Phi_\text{enc}(I,D))
\in \mathbb{R}^{(C-1) \times H \times W}.
$
Note that the network output has $C-1$ channels only, as depth is taken directly from $\Psi$.
Please see the supplement for full details.

\paragraph{Multi-Gaussian prediction.}
\label{s:layered}

While the Gaussians in the model above have the ability to be offset from the corresponding pixel's ray, each Gaussian tends naturally to model the portion of the object that projects onto that pixel.
\citet{szymanowicz24splatter} note that, for individual objects, there is a large number of background pixels that are not associated with any object surface, and these can be repurposed by the model to capture the unobserved parts of the 3D object.
However, this is not the case for scenes, where the goal is to reconstruct every input pixel, and beyond.

Since there are no ``idle'' pixels, it is difficult for the model to repurpose some of the Gaussians to model the 3D scene around occlusions and beyond the image field-of-view.
Hence, we propose to predict a small number $K > 1$
of different Gaussians for each pixel.

Conceptually, given an image $I$ and an estimated depth map $D$, our network predicts a set of shape, location and appearance parameters
$
\mathcal{P}=\{(\sigma_i, \delta_i, \Delta_i, \Sigma_i, c_i)\}_{i=1}^K
$
for every pixel $\bu$, where the depth of the $i$\textsuperscript{th} Gaussian is given by
\begin{equation}
d_i = d + \sum_{j=1}^i \delta_{j},
\end{equation}
where $d = D(\bu)$ is the predicted depth at pixel $\bu$ in depth map $D$ and $\delta_{1} = 0$ is a constant.
Note that since the depth offset $\delta_{i}$ cannot be negative, this ensures that subsequent Gaussian layers are ``behind'' previous ones and encourages the network to model occluded surfaces.
The mean of the $i$\textsuperscript{th} Gaussian is then given by
$
\bmu_i = (u_x \, d_i / f, \, u_y \, d_i / f, \, d_i) + \Delta_i.
$
In practice, we find $K=2$ to be a sufficiently expressive representation.

\paragraph{Reconstructing beyond the border with padding.}

As we show empirically, it is important for the network to be able to model 3D content just outside its field-of-view.
Although multiple Gaussian layers help in this regard, there is a particular need for additional Gaussians near the image border
(\eg, for good new view synthesis when the camera retracts).
To facilitate obtaining such Gaussians, the encoder $\Phi_\text{enc}$ starts with \emph{padding} the input image and depth $(I,D)$ with $P > 0$ pixels on each side, so that the outputs 
$
\Phi_k(I,D) \in \mathbb{R}^{(C-1) \times (H+2P) \times (W+2P)}
$
are larger than the inputs.

\section{Experiments}
\label{sec:manuscript}

We design our experiments to support four key findings, with each section dedicated to one finding.
We begin with the most important result: cross-dataset generalisation---leveraging a monocular depth prediction network and training on a single dataset results in good reconstruction quality on other datasets~(\cref{sec:results_cross}).
Second, we establish that \method serves as an effective representation for single-view 3D reconstruction by comparing against methods specifically designed for this task~(\cref{sec:resuts_single}).
Third, we go as far as to show that the prior learned by single-view \method is as strong as that learned by two-view methods~(\cref{sec:resuts_two}).
Finally, we show via ablation studies how each design choice contributes to performance \method~(\cref{sec:ablations}) and analyse outputs from \method to gain insight into its inner workings.

\subsection{Experiment settings}
\label{sec:results_setting}


\paragraph{Datasets.}
\method is trained only on the large-scale RealEstate10k~\cite{tinghui18stereo} dataset, containing real estate videos from YouTube.
We follow the default training/testing split with 67,477 scenes for training and 7,289 for testing.
%
Once \method is trained, we assess its effectiveness across various datasets, discussed in detail in each section.
For details of the evaluation protocols, see the appendix.

\vspace{-6pt}\paragraph{Metrics.}
For quantitative results,
we report the standard image quality metrics,
including pixel-level PSNR,
patch-level SSIM,
and feature-level LPIPS.

\vspace{-6pt}\paragraph{Compared methods.}
We compare with several state-of-the-art single-view 3D scene reconstruction models,
including LDI~\cite{tulsiani18layer-structured}, Single-View MPI~\cite{tucker20single-view}, SynSin~\cite{wiles20synsin:}, BTS~\cite{wimbauer23behind} and MINE~\cite{li21mine:}.
We include a comparison to our adaptation of Splatter Image~\cite{szymanowicz24splatter} to show that our method is much better suited to general scenes, as well as a comparison to ray-wise unprojection $\mathcal{U}$ of the input colours to the locations predicted by the depth network.
Finally, while not a fair comparison, we also compare with state-of-the-art two-view novel view synthesis methods, including \cite{du23learning}, pixelSplat~\cite{charatan23pixelsplat:}, MVSplat~\cite{chen24mvsplat:}, and latentSplat~\cite{wewer24latentsplat:}.

\vspace{-6pt}\paragraph{Implementation details.}
\method comprises a pre-trained UniDepth~\cite{piccinelli24unidepth:} model, a ResNet50~\cite{he16deep} encoder, alongside multiple depth offset decoders and Gaussian decoders.
The entire model is trained on a single A6000 GPU for 40,000 iterations with batch size 16.
The training is remarkably efficient, completed in one day on a single A6000 GPU.
Given that UniDepth remains frozen during training,
we can expedite the training by pre-extracting depth maps for the entire dataset.
With this, \method can be trained to achieve state-of-the-art quality on a \textit{single A6000 GPU in 16 hours}.

\begin{table}[tb]
  \caption{{\bf Cross-Domain Novel View Synthesis}.
    We evaluate Novel View Synthesis accuracy on datasets not used in training of our method.
    We outperform baselines which were trained on KITTI specifically.
    Here, cross-domain (CD) denotes that the method was not trained on the dataset being evaluated.
  }
  \label{tab:zero-shot}
  \centering
  \footnotesize

  \resizebox{\linewidth}{!}{%
  \begin{tabular}{l cccc cccc}
    \toprule
    {} & \multicolumn{4}{c}{KITTI} & \multicolumn{4}{c}{NYU} \\
    Method & CD & PSNR$\uparrow$ & SSIM$\uparrow$ & LPIPS $\downarrow$ & CD & PSNR $\uparrow$ & SSIM $\uparrow$ & LPIPS $\downarrow$ \\
    \cmidrule{1-1} \cmidrule(l){2-5} \cmidrule(l){6-9}
    LDI~\cite{tulsiani18layer-structured} & \xmark & 16.50 & 0.572 & - & - & - & - \\
    SV-MPI~\cite{tucker20single-view} & \xmark & 19.50 & 0.733 & - & - & - & - & - \\
    BTS~\cite{wimbauer23behind} & \xmark & 20.10 & 0.761 & 0.144 & - & - & - & - \\
    MINE~\cite{li21mine:} & \xmark & 21.90 & \textbf{0.828} & \textbf{0.112} & \cmark & 24.33 & 0.745 & 0.202 \\
    \midrule
    UniDepth w/ $\mathcal{U}$ & \cmark & 20.86 & 0.774 & 0.154 & \cmark & 22.54 & 0.732 & 0.212 \\
    \method (Ours) & \cmark & \textbf{21.96} & 0.826 & 0.132 & \cmark & \textbf{25.45} & \textbf{0.774} & \textbf{0.196} \\
  \bottomrule
  \end{tabular}
  }


\end{table}
\begin{table*}[tb]
  \caption{{\bf In-domain Novel View Synthesis.} 
  Our model shows state-of-the-art in-domain performance on RealEstate10k on small, medium and large baseline ranges.
  }
  \label{tab:re10k_cropped}
  \centering
  \footnotesize
  \setlength{\tabcolsep}{2pt}
  \begin{tabularx}{\linewidth}{@{}l CCC  CCC  CCC @{}}
    \toprule
    {} & \multicolumn{3}{c}{5 frames} & \multicolumn{3}{c}{10 frames} & \multicolumn{3}{c}{$\mathcal{U}[-30,30]$ frames} \\
    Model & PSNR $\uparrow$ & SSIM $\uparrow$ & LPIPS $\downarrow$ & PSNR $\uparrow$ & SSIM $\uparrow$ & LPIPS $\downarrow$ & PSNR $\uparrow$ & SSIM $\uparrow$ & LPIPS $\downarrow$  \\
    \cmidrule{1-1} \cmidrule(l){2-4} \cmidrule(l){5-7} \cmidrule(l){8-10}
    Syn-Sin~\cite{wiles20synsin:} & - & - & - & - & - & - & 22.30 & 0.740 & - \\
    SV-MPI~\cite{tucker20single-view} & 27.10 & 0.870 & - & 24.40 & 0.812 & - & 23.52 & 0.785 & - \\
    BTS~\cite{wimbauer23behind} & - & - & - & - & - & - & 24.00 & 0.755 & 0.194 \\
    Splatter Image~\cite{szymanowicz24splatter} & 28.15 & 0.894 & 0.110 & 25.34 & 0.842 & 0.144 & 24.15 & 0.810 & 0.177 \\ 
    MINE~\cite{li21mine:} & 28.45 & 0.897 & 0.111 & 25.89 & 0.850 & 0.150 & 24.75 & 0.820 & 0.179 \\
    \midrule 
    \method (Ours) & \textbf{28.46} & \textbf{0.899} & \textbf{0.100} & \textbf{25.94} & \textbf{0.857} & \textbf{0.133} & \textbf{24.93} & \textbf{0.833} & \textbf{0.160}  \\
  \bottomrule
  \end{tabularx}
\end{table*}
\begin{figure*}[tb!]
    \centering
    \includegraphics[width=\textwidth]{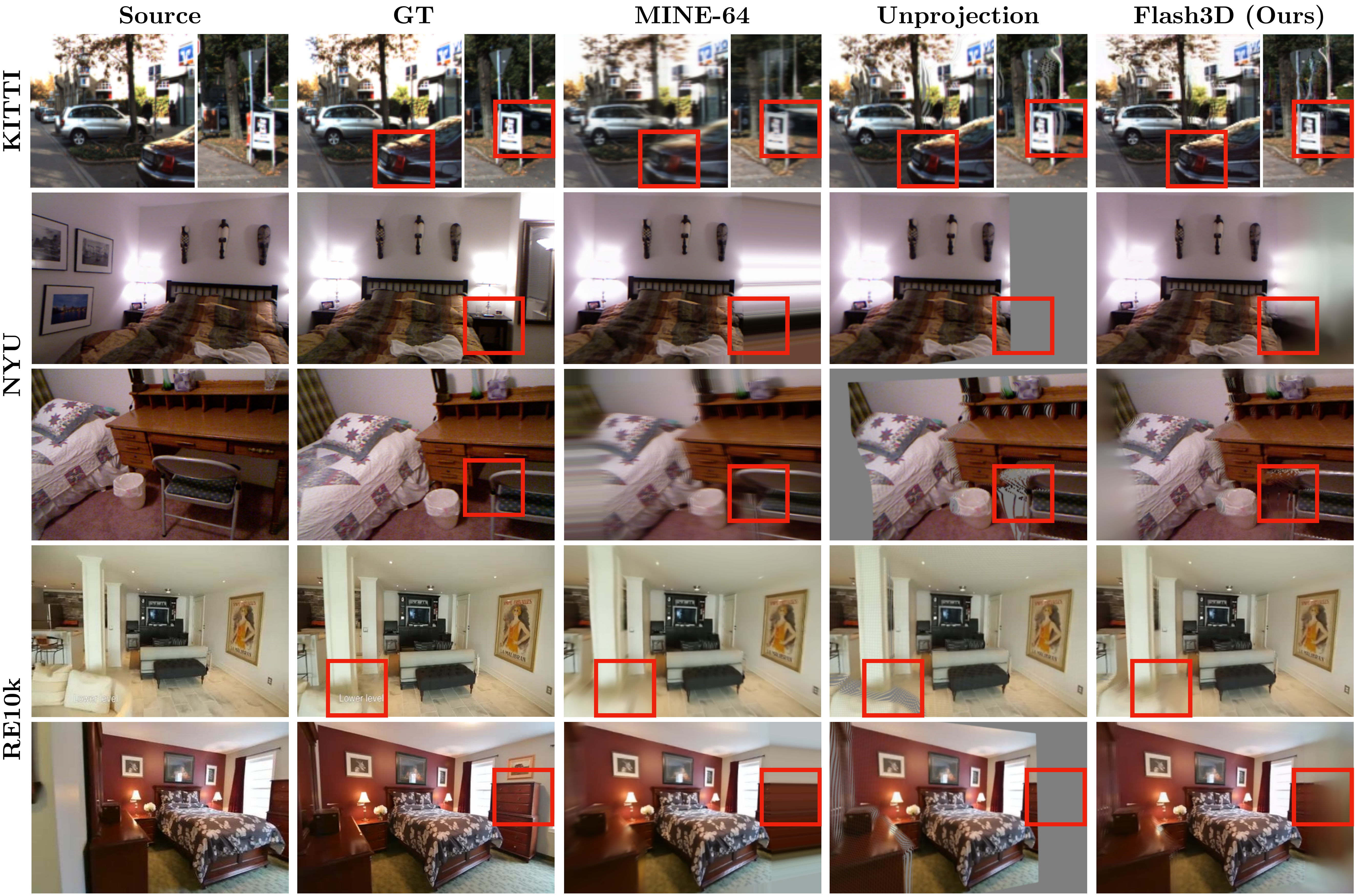}
    \caption{\textbf{Qualitative comparison of monocular reconstruction} on all datasets. \method (Ours, right column) is sharper (top row, car's back) than state-of-the-art MINE~\cite{li21mine:} despite \method not training on KITTI. 
    This is thanks to leveraging a depth predictor which, when used on its own (fourth column), cannot represent occluded regions (third row, fourth row).
    As well as representing occluded regions better than MINE (first row, third row and fourth row), \method also fills in better explanations of regions outside the source camera frustrum (second row, fifth row).}
    \label{fig:in-domain}
\end{figure*}

\subsection{Cross-domain novel view synthesis}
\label{sec:results_cross}

\paragraph{Datasets.}

To evaluate the cross-domain generalisation ability,
we directly evaluate performance on unseen outdoor (KITTI~\cite{geiger12are-we-ready}) and indoor (NYU~\cite{silberman12indoor}) datasets.
For KITTI, we follow standard benchmarks with a well-established protocol for evaluation,
with 1,079 images for testing.
For NYU, we formed a new protocol, with 250 source images for testing (see supp. mat for details).
We evaluate all methods in the same manner.
We verified that Unidepth~\cite{piccinelli24unidepth:} was not trained on these datasets.

To the best of our knowledge, we are the first to report performance on feed-forward cross-domain monocular reconstruction.
We consider two challenging comparisons.
First, we evaluate \method and the current state-of-the-art~\cite{li21mine:} on NYU, an indoor dataset that is similar in nature to RE10k, yet unseen in training.
In~\cref{tab:zero-shot}, we observe that our method performs significantly better on this transfer experiment, despite the domain gap being relatively small.
This suggests that prior works do not generalise as well as our method.
Second, we compare our method on KITTI in \cref{tab:zero-shot}, where it performs on par with the state-of-the-art that was trained on this dataset.
Indeed, \method outperforms the others with respect to PSNR, despite being trained only on an indoor dataset.
This suggests that leveraging a pretrained depth network has allowed our network to learn an extremely strong shape and appearance prior that is even more accurate than learning on this dataset directly.


\begin{table*}[!t]
  \caption{{\bf Ablation Study.}
    Results for ablating different design choices of our method.
  }
  \label{tab:network_ablation}
  \centering
  \footnotesize
  \setlength{\tabcolsep}{1pt}
  \begin{tabularx}{\linewidth}{@{} l  CCC  CCC  CCC  @{}}
    \toprule
     {} & \multicolumn{3}{c}{RE10k -- in-domain} & \multicolumn{3}{c}{NYU -- cross-domain} & \multicolumn{3}{c}{KITTI -- cross-domain} \\
     {} & PSNR $\uparrow$ & SSIM $\uparrow$ & LPIPS $\downarrow$ & PSNR $\uparrow$ & SSIM $\uparrow$ & LPIPS $\downarrow$ & PSNR $\uparrow$ & SSIM $\uparrow$ & LPIPS $\downarrow$ \\
     \cmidrule{1-1} \cmidrule(l){2-4} \cmidrule(l){5-7} \cmidrule(l){8-10}
     \method & \textbf{24.93} & \textbf{0.833} & \textbf{0.160} & \textbf{25.09} & \textbf{0.775} & \textbf{0.182} & \textbf{21.96} & \textbf{0.826} & \textbf{0.132} \\
     w/o depth net, w/ 2nd layer & 23.62 & 0.782 & 0.186 & 23.73 & 0.732 & 0.210 & - & - & - \\ 
     w/o depth net, w/o 2nd layer  & 24.01 & 0.806 & 0.176 & 23.98 & 0.750 & 0.207 & - & - & -  \\
     w/ depth net, w/o 2nd layer & 24.45 & 0.825 & 0.163 & 24.83 & 0.767 & 0.190 & 21.50 & 0.812 & 0.141 \\
     w/ depth net, unproject only & 22.80 & 0.781 & 0.207 & 22.14 & 0.729 & 0.217 & 20.86 & 0.774 & 0.154 \\
  \bottomrule
  \end{tabularx}
\end{table*}

\begin{table}[tb]
  \caption{{\bf Comparison with Two-view Methods}.
    We compare on the split used by pixelSplat \cite{charatan23pixelsplat:} for two-view interpolation and on the split used by latentSplat \cite{wewer24latentsplat:} for extrapolation.
    We take the view closest to the target as the source.
    Our method uses a \emph{single} view and still extrapolates better.
  }
  \label{tab:two_view}
  \centering
  \footnotesize
  \resizebox{\linewidth}{!}{%
  \begin{tabular}{l c ccc ccc}
    \toprule
    {} & Input & \multicolumn{3}{c}{RE10k Interpolation} & \multicolumn{3}{c}{RE10k Extrapolation} \\
    Method & Views & PSNR $\uparrow$ & SSIM $\uparrow$ & LPIPS $\downarrow$ & PSNR $\uparrow$ & SSIM $\uparrow$ & LPIPS $\downarrow$ \\
    \cmidrule{1-2} \cmidrule(l){3-5} \cmidrule(l){6-8}
    Du et al~\cite{du23learning} & 2 & 24.78 & 0.820 & 0.213 & 21.83 & 0.790 & 0.242 \\ 
    pixelSplat~\cite{charatan23pixelsplat:} & 2 & 26.09 & 0.864 & 0.136 & 21.84 & 0.777 & 0.216 \\ 
    latentSplat~\cite{wewer24latentsplat:} & 2 & 23.93 & 0.812 & 0.164 & 22.62 & 0.777 & 0.196 \\
    MVSplat~\cite{chen24mvsplat:} & 2 & \textbf{26.39} & \textbf{0.869} & \textbf{0.128} & 23.04 & 0.813 & \textbf{0.185} \\
    \midrule
    \method (Ours) & \textbf{1} & 23.87 & 0.811 & 0.185 & \textbf{24.10} & \textbf{0.815} & \textbf{0.185} \\ 

  \bottomrule
  \end{tabular}
  }
\vspace{-1em}
\end{table}

\subsection{In-domain novel view synthesis}
\label{sec:resuts_single}


We perform an in-domain evaluation on RealEstate10k~\cite{tinghui18stereo}, following the same protocol as prior works~\cite{li21mine:}.
We evaluate the quality of zero-shot reconstruction and compare performance on an in-domain dataset, RealEstate10k. 
We evaluate the quality of reconstruction through novel view synthesis metrics as that is the only ground-truth data available in this dataset.
RealEstate10k evaluates the quality of reconstructions at different distances between the source and the target, as a smaller distance makes the task easier.
In~\cref{tab:re10k_cropped}, we observe that we achieve state-of-the-art results on this mature benchmark across all distances between the source and the target.
Further analysis in~\cref{fig:in-domain} reveals that our method's reconstructions are sharper and more accurate than prior state-of-the-art~\cite{li21mine:}, despite being trained on an order of magnitude fewer GPUs (1 vs. 64).

\subsection{Comparison to few-view novel view synthesis}
\label{sec:resuts_two}

\paragraph{Datasets.}

To further evaluate the effectiveness of \method, we conducted assessments using the pixelSplat~\cite{charatan23pixelsplat:} split for interpolation and the latentSplat~\cite{wewer24latentsplat:} split for extrapolation.

Unlike existing two-view methods that typically assess interpolation between two source views,
\method consistently performs extrapolation from a single view.
The results are reported in \cref{tab:two_view}.
Here, \method cannot outperform two-view approaches on the interpolation task, due to receiving less information.
However, \method surpasses all previous state-of-the-art two-view methods at view extrapolation.
This highlights the utility of the multi-layer Gaussian representation of our approach at capturing and modelling unseen areas.

\begin{figure*}[tb!]
    \centering
    \includegraphics[width=\textwidth]{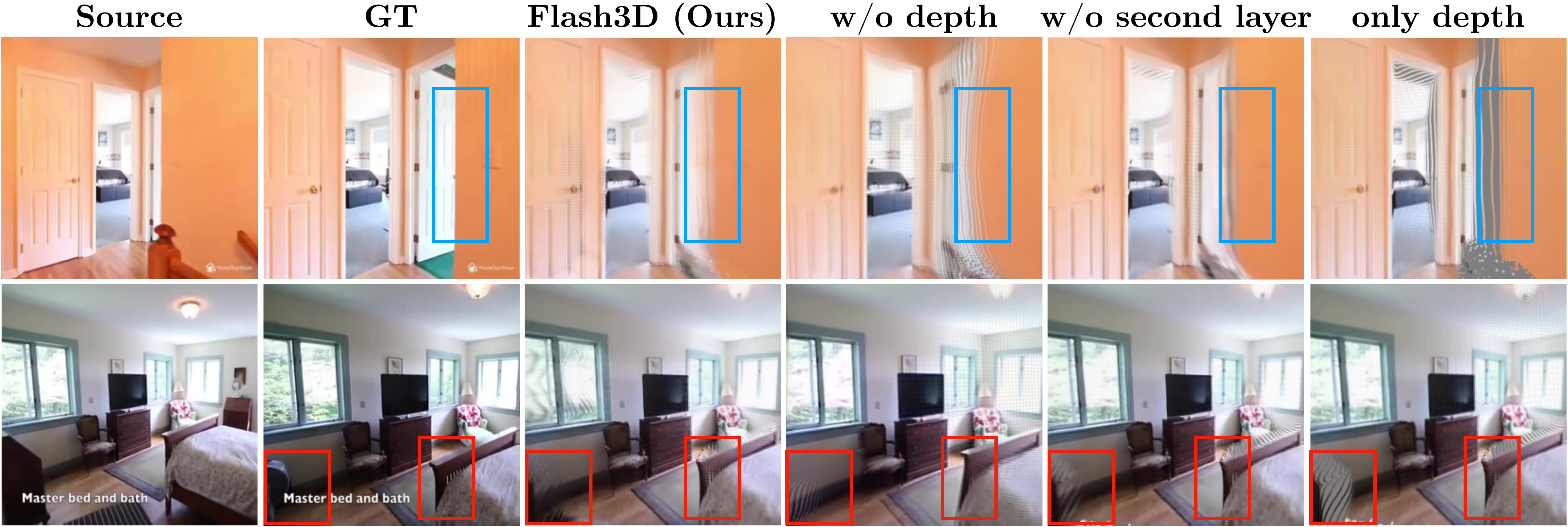}
    \caption{\textbf{Ablation.}
    We show how \method degrades when components are removed.
    Removing the depth network (4\textsuperscript{th} column) results in incorrect geometry (orange wall, corner of the bed).
    Using only one layer of Gaussians (5\textsuperscript{th}) results in holes in renderings due to disocclusions (orange wall, area behind cabinet), although they are not as bad as when simply using depth unprojection (rightmost).}\vspace{-0.5em}
    \label{fig:ablation}
\end{figure*}
\begin{figure*}
    \centering
    \includegraphics[width=\textwidth]{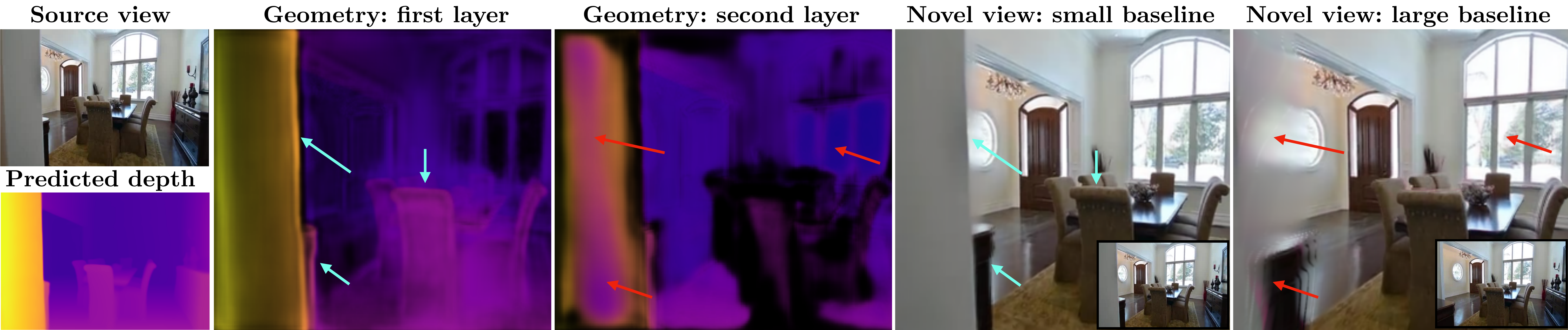}
    
    \caption{\textbf{Analysis of layered Gaussians.} The first layer of Gaussians (second column) represents visible parts where the depth prediction can be used (blue arrows). 
    The second layer (third column) represents the remaining parts of the scene (red arrows): occluded regions (wall, cabinet) and regions where depth prediction is unreliable (windows).
    Combining the two leads to sharp geometries at small baselines (fourth column) and reasonable reconstructions at large baselines (right column).}\vspace{-0.5em}
    \label{fig:analysis}
\end{figure*}

\subsection{Ablation study and analysis}
\label{sec:ablations}

\paragraph{Ablation study.} 
We ablate our method for in-domain and cross-domain settings in~\cref{tab:network_ablation}, focusing on the following questions.
\textbf{Q1:} 
Is leveraging a monocular depth predictor useful in the task of reconstructing appearance and geometry of scenes?
\textbf{Q2:} 
If yes, is it sufficient on its own, \ie, is learning shape and appearance parameters necessary for scene reconstruction with 3D Gaussians?

\emph{Importance of depth predictor.}
We remove the pretrained depth network that predicts depth $D$, instead estimating it jointly with all other parameters.
First, in~\cref{tab:network_ablation} we observe that this leads to a significant drop in performance compared to \method, 
indicating that the depth network contains important cues that had already been learned. 
Moreover, the third row of~\cref{tab:network_ablation} indicates that without the depth network, 2 layers of Gaussians per pixel performs \textit{worse} than using just one layer. 
We hypothesise that the depth network plays an important role in avoiding local optima that were reported to limit the learning capabilities of primitive-based methods~\cite{charatan23pixelsplat:}.
Qualitatively, the fourth column in~\cref{fig:ablation} illustrates that removing the depth network makes it challenging to learn accurate geometries of walls (orange wall is bent) and object boundaries (bed has an incorrect shape).

\emph{Importance of extending beyond depth prediction.}
Here we remove the learned parts of our network.
First, we use only one layer of Gaussians and predict parameters $\mathcal{P}_1$ corresponding to the depth $D$ predicted by the pre-trained depth network.
This also results in a drop in performance in~\cref{tab:network_ablation}.
We then go further and remove the network that predicts $\mathcal{P}_{1}$, removing learning altogether.
Novel views are simply rendered from source view colours backprojected with the depths $D$.
This understandably drops the performance even further.
In~\cref{fig:ablation} we observe that these drops are due to the 1-layer method not representing occluded parts of the scene.
The last column in~\cref{fig:ablation} illustrates that the holes are significant when using only depth unprojection, and can be partially mitigated when learning shapes $\mathcal{P}_1$ due to the network being able to stretch the Gaussians at depth discontinuities.


\paragraph{Analysis.}
~\cref{fig:analysis} analyses the contribution of each Gaussian layer to a full reconstruction of the scene.
We visualise the depth of each of the layers, $D$ and $D+\delta_2$, multiplied by the opacity $\sigma_1,\sigma_2$ of the corresponding Gaussians, illustrating how much they are used when rendering the scene.
In~\cref{fig:analysis}, the more saturated the colour, the more opaque the Gaussian (black is fully transparent).
We observe that the first layer has the most opaque Gaussians at object boundaries (wall, cabinet) and at complicated geometries (chair), indicating that these are the regions where the depth prediction network is the most useful.
This is further supported by \cref{fig:ablation} where removing the depth network impaired reconstruction in exactly the same regions.
Leveraging the depth network at object boundaries results in crisp, accurate geometries at small baselines (fourth column).
Interestingly, the network ignores the depth prediction for windows, which are consistently incorrect.
Next, we analyse where the second layer of Gaussians places its predictions.
In the third column of~\cref{fig:analysis},
where the network observes a wall, the second layer of Gaussians is placed at larger depth.
These Gaussians are observed when the camera motion (baseline) is large, and are shown to have a reasonable appearance in~\cref{fig:analysis} (right column).
The last column in~\cref{fig:analysis} additionally reveals a limitation of our method.
It is a deterministic, regressive model of structure and appearance, and thus produces blurry renderings in presence of ambiguity: when baselines are very large, in occluded regions or when camera moves backward.
Blurriness could be reduced with additional losses (perceptual~\cite{zhang18the-unreasonable} or adversarial~\cite{goodfellow14generative}).
Alternatively, our method could be incorporated as conditioning within a framework similar to~\cite{chan23generative} or as the reconstructor in a diffusion-based feed-forward 3D generation framework~\cite{szymanowicz23viewset,tewari23diffusion}.

\section{Conclusion}
\label{sec:conclusions}

We presented \method, a model that can be trained in just
16h on a single GPU to achieve state-of-the-art results for monocular scene reconstruction.
Our formulation allows using a monocular depth estimator as a foundation for full 3D scene reconstruction.
As a consequence, the model generalizes very well:
it outperforms prior works even when not trained specifically on the target dataset as them.
Analyses reveal the interaction mechanisms between the pretrained network and the learned modules, and ablations verify the importance of each component.

\vspace{-1em}
\paragraph{Acknowledgements.} The authors acknowledge the generous support of the Royal Academy of Engineering (RF\textbackslash 201819\textbackslash 18\textbackslash 163), EPSRC SYN3D Grant EP/Z001811/1, EPSRC Doctoral Training Partnerships Scholarship (DTP) EP/R513295/1 and the Oxford-Ashton Scholarship.
For the purpose of Open Access, the author has applied a CC BY public copyright licence to any Author Accepted Manuscript (AAM) version arising from this submission (\url{https://openaccess.ox.ac.uk/rights-retention/}).

{
    \small
    \bibliographystyle{ieeenat_fullname}
    \bibliography{main,vedaldi_general,vedaldi_specific}
}

\begin{figure*}[!t]
    \centering
    \includegraphics[width=\textwidth]{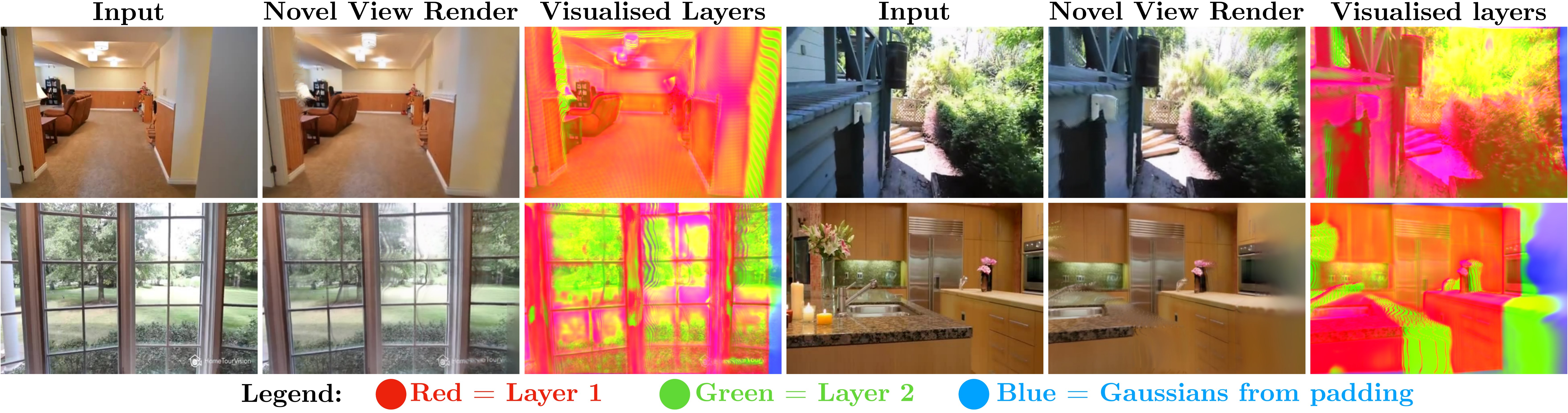}
    \caption{\textbf{Analysis of Gaussian allocation.} 
    Gaussians from the first layer (red) are allocated in visible parts, from the second layer (green) in occluded regions (top row, bottom right) and on windows (bottom left) and Gaussians from the padding region (blue) are revealed when camera reveals regions that were not present in the frustrum of the input camera.
    }
    \label{fig:analysis_colors}
\end{figure*}

\section{Dataset details}

\paragraph{RealEstate10k.}
We download the videos from provided links, resulting in above 65,000 videos, as well as the provided camera pose trajectories.
Using the provided cameras, we run sparse point cloud reconstruction with COLMAP~\cite{schonberger16structure-from-motion}.
We use the test split provided by MINE, and following prior work we evaluate PSNR on novel frames which are 5 and 10 frames ahead of the source frame.
In addition, we evaluate on a random frame sampled from an interval of $\pm30$ frames.
We use the same frames as~\cite{li21mine:} did for their evaluation.
As a result, we evaluate on 3205 frames.
We reproduced the results from~\cite{li21mine:} using their released checkpoint with the common protocol of cropping 5\% of the image around the border, achieving scores similar to those presented in the original paper.
We confirmed with authors of BTS~\cite{wimbauer23behind} that this is the commonly used protocol.
We do our training and testing at $256 \times 384$ resolution.

\paragraph{NYUv2.}
We form a benchmark that is similar in nature to RealEstate10k in that it shows indoor scenes, but is visually radically different.
We download 80 raw sequences of NYUv2~\cite{silberman12indoor} and run COLMAP~\cite{schonberger16structure-from-motion} on them to recover camera pose trajectories.
On each video we sample 3 random souce frames and use a random frame uniformly sampled within $\pm30$ frames from the source frame, mirroring the protocol of RealEstate10k.
We undistort images, and rescale to $256 \times 384$ resolution. 

\paragraph{KITTI}
We evaluate on the Tulsiani test split~\cite{tulsiani18layer-structured} of the KITTI~\cite{geiger13vision} dataset.
The cameras in the KITTI dataset are in metric scale, our network works directly with the provided cameras and scenes without additional preprocessing.
For evaluation, following prior work~\cite{wimbauer23behind,li21mine:} we crop the outer 5\% of the images.


\section{Baselines and competing methods}

\subsection{Depth unprojection}
A crucial baseline in our experiments is measuring performance of monocular depth prediction for monocular Novel View Synthesis.
In this baseline, we place isotropic 3D Gaussians with fixed opacity without view-dependent effects (i.e. a point cloud with soft point boundaries) at the depths predicted by the monocular depth predictor.
We set the Gaussian colours to be a scaled copy from the input view so that $c_{G} = \alpha c_{RGB}$ and we initialise $\alpha=1.0$.
We initialise Gaussian opacity to be $\sigma = \text{sigmoid}(\sigma_{0})$, with $\sigma_0=4.0$, i.e., almost opaque.
We test two variants of setting the scale of Gaussians: (1) one where Gaussians have a fixed scale $s = \exp{s_0}$ with $s_0=-4.5$, and (2) one where the radius is proportional to depth from camera, allowing the Gaussians to fit inside the ray cast from the pixel: $s = \exp{s_0} d / d_0$, where $d$ is metric depth output from UniDepth, $d_0=10.0$ and $s_0=-4.5$.
Next, while we determined $\alpha=1.0$, $s_0=-4.5$ and $\sigma_0=4.0$ to be reasonable initialisations, they might not correspond to the highest quality of Novel View Synthesis. 
Thus, we run gradient-based optimisation of the parameters of this baseline, optimising $\alpha, s_0, \sigma_0$ to minimise the photometric loss in the source view and 3 novel views (identical to our final model) on the training set.
We train these models for $5,000$ iterations and choose the one with the best performance on validation split.
Finally, we evaluate the model with the best $\alpha, s_0, \sigma_0$ on the test split and report the metrics.

\begin{table*}[h]
  \caption{{\bf Depth Unprojection Baseline}.
    We fit hyperparameters of the depth unprojection model via gradient-based optimisation.
    We try two variants: one with fixed-size Gaussians and one where the Gaussian scale is increased proportionally to depth.
    Top two rows are before correcting depth-wise unprojection to be from pixel centers instead of pixel corners.
    All measured with croppint.
  }
  \label{tab:depth_unprojection_tune}
  \centering
  \resizebox{0.99\linewidth}{!}{%

  \begin{tabular}{@{} l l ccc  ccc  ccc @{}}
    \toprule
    {} & {} & \multicolumn{3}{c}{5 frames} & \multicolumn{3}{c}{10 frames} & \multicolumn{3}{c}{random frame} \\
    Model & Backbone & PSNR $\uparrow$ & SSIM $\uparrow$ & LPIPS $\downarrow$ & PSNR $\uparrow$ & SSIM $\uparrow$ & LPIPS $\downarrow$ & PSNR $\uparrow$ & SSIM $\uparrow$ & LPIPS $\downarrow$  \\
    \cmidrule{1-2}\cmidrule(l){3-5}\cmidrule(l){6-8}\cmidrule(l){9-11}
    Fixed size & ConvNeXT-L & 26.47 & 0.864 & 0.120 & 24.08 & 0.808 & 0.173 & 22.60 & 0.774 & 0.211 \\
    Fixed size & ViT-L & 26.62 & \textbf{0.867} & \textbf{0.120} & 24.25 & \textbf{0.814} & \textbf{0.172} & 22.78 & \textbf{0.781} & 0.209 \\
    \midrule
    Depth-dependent & ConvNeXT-L & 26.49 & 0.861 & 0.124 & 24.10 & 0.806 & 0.175 & 22.61 & 0.774 & 0.209 \\
    Depth-dependent & ViT-L & \textbf{26.65} & 0.864 & 0.123 & \textbf{24.29} & 0.812 & 0.173 & \textbf{22.80} & \textbf{0.781} & \textbf{0.207} \\
  \bottomrule
  \end{tabular}
  }
\end{table*}

\subsection{Splatter Image}

We implemented the Splatter Image baseline using the same U-Net convolutional neural network with a ResNet-50 backbone as our own method for a fair comparison. We trained it on two NVIDIA A6000 GPUs for a total of $350,000$ steps, an order of magnitude more than our proposed \method.
Training took 6GPU days, same as reported in~\cite{szymanowicz24splatter}.

\subsection{MINE}
MINE~\cite{li21mine:} only provided model weights but no inference and evaluation code on RealEstate10K dataset, hence we re-run the inference and evaluation for reproducibility. 
The results match those reported in ~\cite{li21mine:}.
We use the $N=64$ model since that is the best one made available by the authors.
For evaluation on NYU we use the model trained on Re10k, identically to our method.

\subsection{Two-view methods}
When comparing to two-view methods, we ought to choose one of them as our source view. 
For any method, the most indicative factor of performance on a target frame is the baseline to the source frame.
We run this comparison on $256\times256$ without border-cropping for being comparable.

\begin{table*}[h]
  \caption{{\bf Ablation Study for Depth Decoder Architectures.}
    Here, we ablate the probabilistic depth as in pixelSplat~\cite{charatan23pixelsplat:},
    but only for the $K > 1$ of Gaussians.
    $-K$ means $K$ Gaussians per-pixel.
    Here, cross-domain (CD) denotes that the method was not trained on the dataset being evaluated.
  }
  \label{tab:abla_depth}
  \centering
  \footnotesize
  \begin{tabularx}{\linewidth}{@{}l cCCC cCCC @{}}
    \toprule
    {} & \multicolumn{4}{c}{KITTI} & \multicolumn{4}{c}{NYU} \\
    Method & CD & PSNR $\uparrow$ & SSIM $\uparrow$ & LPIPS $\downarrow$ & CD & PSNR $\uparrow$ & SSIM $\uparrow$ & LPIPS $\downarrow$ \\
    \cmidrule{1-1} \cmidrule(l){2-5} \cmidrule(l){6-9}
   \method (Discrete)-2 & \cmark & 21.35 & 0.805 & 0.153 & \cmark & 24.52 & 0.763 & 0.200 \\
   \method (Discrete)-3 & \cmark & 21.50 & 0.814 & 0.136 & \cmark & 24.84 & 0.772 & 0.189 \\
    \midrule
    \method (Ours)-2 & \cmark & \textbf{21.96} & \textbf{0.826} & \textbf{0.132} & \cmark & \textbf{25.09} & \textbf{0.775} & \textbf{0.182} \\
  \bottomrule
  \end{tabularx}
\end{table*}

\subsection{Probability distribution of Gaussian}
An alternative approach to the multiple Gaussians is to predict depth probabilities as in pixelSplat~\cite{charatan23pixelsplat:}.
However, without the estimated depth from the pre-trained depth predictor, the coverage speed is very slow, and the performance is worse in our monocular setting.
For a fair comparison, we ablate only on other Gaussian layers, \ie $K > 1$ of Gaussians.
The results are reported in \cref{tab:abla_depth}.
The continuous depth offset outperforms the depth probabilities design in pixelSplat.

\begin{table*}[h]
  \caption{{\bf Ablations on different depth models}.
    We fit hyperparameters of the depth unprojection model via gradient-based optimisation.
    We try two variants: one with fixed-size Gaussians and one where the Gaussian scale is increased proportionally to depth.
    Top two rows are before correcting depth-wise unprojection to be from pixel centers instead of pixel corners.
    All measured with croppint.
  }
  \label{tab:depth_ablation}
  \centering
  \resizebox{0.99\linewidth}{!}{%

  \begin{tabular}{@{} l ccc  ccc  ccc @{}}
    \toprule
    {} & \multicolumn{3}{c}{5 frames} & \multicolumn{3}{c}{10 frames} & \multicolumn{3}{c}{random frame} \\
    Model & PSNR $\uparrow$ & SSIM $\uparrow$ & LPIPS $\downarrow$ & PSNR $\uparrow$ & SSIM $\uparrow$ & LPIPS $\downarrow$ & PSNR $\uparrow$ & SSIM $\uparrow$ & LPIPS $\downarrow$  \\
    
    \cmidrule{1-1}\cmidrule(l){2-4}\cmidrule(l){5-7}\cmidrule(l){8-10}
    Unidepth V1 & \textbf{28.46} & \textbf{0.899} & \textbf{0.100} & \textbf{25.94} & \textbf{0.857} & \textbf{0.133} & \textbf{24.93} & \textbf{0.833} & \textbf{0.160} \\
    DepthAnything V2 & 28.31 & 0.895 & 0.101 & 25.79 & 0.849 & 0.136 & 24.49 & 0.823 & 0.165 \\

    Metric3D V2 & 28.00 & 0.893 & 0.107 & 25.62 & 0.852 & 0.140 & 24.55 & 0.826 & 0.167 \\

  \bottomrule
  \end{tabular}
  }
\end{table*}
\subsection{Off-the-Shelf Depth Models}
We also assess the effect of different monocular depth estimation methods.
We first evaluate the recent DepthAnthing V2~\cite{yang2024depthanything,depth_anything_v2} model,
which provides better details for depth prediction.
However, since their metric depth is either trained only for indoor scenes (Hypersim) or outdoor scenes (KITTI),
we used the indoor checkpoints as the metric depth.
As shown in the ~\cref{tab:depth_ablation},
our framework also achieves comparable results using depths from DepthAnthing V2,
without adjusting any hyper-parameters. Secondly, we evaluated our method using another recent Metric3D V2 monocular depth estimation model~\cite{hu2024metric3dv2}. Similarly, the results are comparable to our main model reaffirming the choice of Unidepth~\cite{piccinelli2024unidepth} as the backbone in our method.

\section{Implementation details}

\subsection{Architecture}

We base our convolutional network on a ResNet-50~\cite{he16deep} backbone and implement a U-Net~\cite{ronneberger15u-net:} encoder-decoder as in~\cite{godard19digging}. Specifically, a single ResNet encoder is shared by a multiple decoders, one for each layer of appearance parameters as well as depth offset decoders, barring the offset decoder for the first layer as we obtain depth values directly from a pre-trained model.

\subsection{Optimisation}

We define the photometric loss following \cite{godard16unsupervised} as a weighted sum of $L_1$ and SSIM~\cite{wang04bimage} terms:

\begin{equation}
\mathcal{L} = \| \hat{J} - J\| + \alpha \operatorname{SSIM}(\hat{J}, J)
\end{equation}

Unlike previous works~\cite{li21mine:,tucker20single-view}, we do not use sparse depth supervision.

where $J$ is a target image, $\hat{J}$ is a rendering, and $\alpha=0.85$. We optimise the network with Adam~\cite{kingma15adam:} with batch size $16$ and a learning rate of $0.0001$ for a total of $40,000$ training steps.

\subsection{Scale alignment}

Camera poses are typically estimated with COLMAP.
These camera poses are in an arbitrary scale in each scene.
Following prior work, we align the scale of the COLMAP cameras to those estimated by our network using the scale factor computation from~\cite{tucker20single-view}.
However, if there are outliers in depth estimation (both in our method and baselines), they will impact the scale estimation.
As a result, there might be mismatch between the scene reconstruction scale and the scale of camera poses from which novel views are rendered.
In consequence, the rendered novel views can be shifted compared to ground truth, which does not significantly impact LPIPS but it does affect PSNR.
Thus, at test-time we run scale alignment with RANSAC.
We do the same for MINE when evaluating it on the transfer dataset, NYU, since the accuracy of its depth prediction deteriorates in this unseen dataset.
When estimating scale we thus use the RANSAC scheme with sample size of 5, $1,000$ iterations and threshold $0.1$.





\section{Limitations}
\label{sec:limitations}

A primary limitation of the proposed approach is due to it being a deterministic, regressive model.
This incentivises it to generate blurry renderings in presence of ambiguity, such as when baselines are very large, in occluded regions or when camera moves backward.

Another limitation is that not all occluded surfaces are captured by the reconstructor: the reconstructed 3D models still have some holes.
While many of these regions are filled in, some are missed, even when multiple Gaussians are predicted.

Finally, failures in the pre-trained depth estimator are likely to lead to failures in our scene reconstructions, especially if the estimated depth is over-estimated.
This is due to the non-negativity of our depth offsets, which therefore cannot recover scene structure closer to the camera than the surface estimated by the pre-trained depth estimator.
This makes the model dependent on the quality of a third-party model within the domain of use at inference time.

\section{Broader impacts}
\label{sec:impacts}

This work, on monocular scene reconstruction, has potential positive and negative social impacts.
On the positive side, the approach significantly reduces the compute and time resources needed to acquire 3D assets in-the-wild, opening the door to consumer applications with positive impacts.
For example, the ability to quickly reconstruct one's house to facilitate its sale; the ability to digitally preserve artefacts and sites of cultural heritage; and uses in safe autonomous driving.

On the negative side, this technology has the potential to be used for malicious purposes, such as illegal or unethical tracking and surveillance, or be invasive of someone's privacy, for example by reconstructing their body without their consent.
In addition, incorrect predictions may cause harm if used in applications like autonomous driving and robotics, where mis-estimated 3D structures could lead to crashes or suboptimal performance.

\end{document}